\documentclass{article}

\usepackage[nonatbib,preprint]{neurips_2023}

\usepackage[utf8]{inputenc} 
\usepackage[T1]{fontenc}    
\usepackage[breaklinks]{hyperref}       
\usepackage{url}            
\usepackage{booktabs}       
\usepackage{amsfonts}       
\usepackage{nicefrac}       
\usepackage{microtype}      
\usepackage{xcolor}         

\hypersetup{
   breaklinks=true,   
}
        
\title{Concept Alignment}

\author{%
    Sunayana Rane \thanks{ Equal contribution.} 
     \\
  Department of Computer Science\\
  Princeton University\\
  Princeton, NJ 08540 \\
  \texttt{srane@princeton.edu} \\
  \And
        Polyphony J. Bruna \footnotemark[1]
     \\
  Department of Cognitive and Information Sciences \\
  University of California Merced\\
  Merced, CA 95343 \\ 
  \texttt{pbruna@ucmerced.edu} \\
\And
  Ilia Sucholutsky \\ 
  Department of Computer Science \\ 
  Princeton University \\
  Princeton NJ, 08540 \\
  \texttt{is2961@princeton.edu} \\ 
  \And
  Christopher Kello \\ 
  Department of Cognitive and Information Sciences \\ 
  University of California Merced \\
  Merced, CA 95343 \\
  \texttt{ckello@ucmerced.edu} \\ 
  \And
  Thomas L. Griffiths \\ 
  Department of Psychology \\
  Department of Computer Science\\
  Princeton University \\
  Princeton NJ, 08540 \\
  \texttt{tomg@princeton.edu}
}

\begin{document}

\maketitle

\begin{abstract}
  Discussion of AI alignment (alignment between humans and AI systems) has focused on \textit{value alignment}, broadly referring to creating AI systems that share human values. We argue that before we can even attempt to align values, it is imperative that AI systems and humans align the concepts they use to understand the world. We integrate ideas from philosophy, cognitive science, and deep learning to explain the need for \textit{concept alignment}, not just value alignment, between humans and machines. We summarize existing accounts of how humans and machines currently learn concepts, and we outline opportunities and challenges in the path towards shared concepts. Finally, we explain how we can leverage the tools already being developed in cognitive science and AI research to accelerate progress towards concept alignment.
\end{abstract}


\section{Introduction}

Alignment of AI with humans is a critically important vision, and a complex technical and societal challenge \cite{christian2020alignment}. Most discussion of AI alignment focuses on \textit{value alignment}, which has been defined broadly and often in the public discourse, for example as ``AI that does what it morally ought to do, as defined by the individual or society,'' \cite{gabriel2020artificial} or ``AI systems [that] act in accordance with our ethical principles and objectives'' \cite{su}. What this means, however, is poorly delineated even for those creating these definitions. Value alignment is discussed more frequently in news articles, blogs, and opinion pieces \cite{newyorker, su} than it is in the academic literature (although see \cite{brown2021value, hadfield2016cooperative, hadfield2017inverse, hendrycks2020aligning}). This important goal is difficult to break down into a tractable path to implementation. Any attempt at a general, all-encompassing solution is immediately riddled with thorny and seemingly insurmountable questions: What exactly is a value for an AI? To whose values should AI be aligned? How do these values change in a given context? Does the AI even understand what it means for a human to have values?

These are difficult questions that require a deeper look at AI alignment. It is important to remember that even \textit{humans} disagree strongly, and often violently, on values. That is one reason why we have democratic governments that reflect a changing set of societal values, and a legal system to periodically redefine the practical boundaries of how those values translate to the real world, with all their edge cases and exceptions, case by case. We also have values as defined in religious systems, economic systems, and even educational systems, and we all have a societal role in shaping our world such that these values are fairly balanced between opposing factions whose values clash. This process is complicated and the results are always imperfect, but we do at least have a process, and it is important that the process be a democratic one that is accessible to everyone, and enriched by a broad and diverse range of perspectives. However, creating value alignment among humans is arguably considerably easier than between humans and machines for one simple reason: humans already have a lot in common in how they conceive of the world around them.

In this paper we argue that before we can even attempt human-AI value alignment, it is important to pursue human-AI alignment at the conceptual level, such that AI systems understand the world in terms of the same concepts humans use to understand the world. We call this {\em concept alignment}. 
The importance of concept alignment can be recognized by understanding where it fails among humans. The process of scientific discovery is awash with the confrontation and resolution of \textit{incommensurate} ways of viewing the world: conceptual frameworks that oppose one another to the extent that it becomes impossible to draw meaningful comparisons between them or to assess the one by the standards of the other. They are more than simply different, they cut across the same conceptual space in different ways \cite{Taylor-1982}, making it impossible to agree on concepts let alone values.

For example, an Aristotelian physicist and a Newtonian physicist, although sharing the term ``motion,'' could not engage in a productive dialogue on the topic because Aristotelian physics is grounded in explaining changes of quality whereas Newtonian physics is grounded in explaining projectile mechanics. These two systems of thought are incommensurable because it is not possible to evaluate the claims of one system within the paradigm of the other, which make radically different ontological assumptions about the world \cite{Kuhn-1978}. Our scientific language is, in fact, littered with the remnants of past worldviews -- lest we forget that ``atom,'' from the ancient Greek ``atomos,'' means \textit{indivisible} (due to the long-held belief that atoms were the fundamental unit of matter) and that ``planet,'' from the ancient Greek ``planetes,'' means \textit{wanderer} (due to their apparent motion against the backdrop of the supposedly fixed stars). Two physicists with such different worldviews would face significant challenges in agreeing on a set of values to guide their science.\footnote{For example, they might disagree on their scientific goals (i.e., which scientific questions are worth pursing).}

We also see significant differences in concepts between children and adults, such as how children understand liquids before acquiring the correct conservation principle \cite{piaget-1967}. When faced with two glass containers, one short and stout and the other tall and thin, filled with the same quantity of liquid, young children will believe that the taller glass contains more liquid. This is despite the fact that the children have witnessed the same quantity of liquid being poured into the two containers. Two people, a young child and an adult, will believe two contradictory facts (that the taller glass contains more liquid / that the amount of liquid contained by the glasses is the same) about the same set of circumstances due to a fundamental conceptual difference -- the adult has access to the concept of volume while the child does not. As a consequence, it might be hard for an adult and a child to agree on the fair allocation of juice, let alone more complex problems of value alignment.

 In the context of contemporary AI advancements, it is unclear to what extent our current AI technologies' conceptual understandings of the world we inhabit are similar to our own. We have many reasons to believe that there are substantial, potentially catastrophic differences. For example, AI classifying black people as non-human primates \cite{raceai}, a self-driving Uber that killed a jaywalking pedestrian because it didn't understand the concept of crossing the street without using a crosswalk \cite{uber}, and an AI that learns the word label ``wolf'' to mean the snow in the background of images of wolves \cite{ribeiro2016should}. Despite using the same language labels like ``wolf'' and ``pedestrian,'' the AI is fundamentally misaligned with our understanding of the concepts those words represent. How can we begin to make progress towards value alignment when the concepts we use in our discussions about it are often so grossly misaligned with the concepts the AI system is using to represent the world?

To inform the debate on how to achieve value alignment between humans and AI, we draw from philosophy to motivate why concept alignment is necessary for safer and more effective AI, and we explain how the fields of cognitive science and AI research are already developing the tools needed to accelerate progress towards concept alignment.


\section{Goals for concept alignment}

Achieving alignment is difficult, even between humans. So what is good enough? What should we aim for in concept alignment research?

One very concrete goal is: 
\begin{quote}
Concept alignment will lead to natural language communication with AI systems that is functionally equivalent to natural language communication among humans.
\end{quote}

It's commonly understood, especially in the age of deep learning, that AI outputs are, to humans, often seemingly ``weird,'' and outside the distribution of what one would expect of human behavior for the same task. Often an AI will find a solution to a simple problem \cite{zhong2023clock} in a way that is outside the range of known solutions that humans would produce. AlphaGo \cite{silver2016mastering} famously did so very successfully, defeating one of the world's best players in a game of Go by using a move the player said no human would have made \cite{alphago-wired}. This type of behavior may be highly desirable when we are seeking innovation and discovery. However, for cases in which AI is to be integrated into society, such as self-driving cars that make the type of mistakes human drivers wouldn't normally make, AI that behaves in unpredictable and uninterpretable ways is often catastrophic and disturbing. One goal in concept alignment research is to produce AI that will be easier to interact with because we will understand it and it will understand us. The natural language that we use to communicate with it will have a sufficiently shared meaning for it to be functionally equivalent to communicating and collaborating with a human, and this means fewer non-human-like, ``weird'' mistakes.

\section{The challenge for shared concepts}

Is an AI's use of natural language (like English) enough to convince us that the meaning behind the words it uses is the same? When ChatGPT \cite{chatgpt} produces words, do you trust that its understanding of those words is the same as your own? Is the meaning behind each word in ChatGPT's English the same as the meaning behind each word you use when chatting with a friend or colleague? Can you make the same assumptions about the meaning of the AI's words as you do about another human?

The notion that language underspecifies the world was formalized in 20th century epistemic critiques in analytic philosophy, spearheaded by W. V. O. Quine. Quine's ``naturalized epistemology'' offered a rebuttal to rationalist, Cartesian foundationalist epistemology, which grounded human knowledge in \textit{a priori} truths and epistemic certainty.\footnote{For Quine's argument against the possibility of an \textit{a priori} foundation for knowledge, see \cite{Quine-1951}} Instead, Quine argues that knowledge is ultimately grounded in the relationship between and development of scientific theories and the empirical evidence that supports them.\footnote{Coherentist approaches like this were also defended by the philosopher of science Otto Neurath \cite{Neurath-1983} and by the epistemologist Laurence BonJour \cite{BonJour-1985}, among others.}

An upshot of grounding knowledge in empirical evidence for Quine is his thesis of semantic indeterminacy, the thesis that translation, meaning, and reference are indeterminate. A primary argument for semantic indeterminacy comes from Quine's argument for the inscrutability of reference (also known as ``ontological relativity'') \cite{Quine-1968}. This argument has since been picked up by fields such as cognitive science, where it is often better known as the ``gavagai argument.''

The gavagai argument asks us to imagine a linguist tasked with translating an unknown foreign language is faced with translating the term ``gavagai'' after witnessing a native speaker utter the term whilst pointing into the distance wherein there sits a rabbit among the grass. Quine argues that there is an infinite number of possible referents of the term ``gavagai'' (e.g., ``rabbit,'' ``rabbit (temporal) stage,'' ``undetached rabbit part''), all of which perfectly describe any evidence that the linguist may gather from the behavior of the native speaker. The argument illustrates that language fundamentally underdetermines meaning.

Quine's indeterminacy theses raise a historic, philosophical challenge to the simple notion that mere participation in a language game affords one access to the concepts that one's interlocutor intends.  We argue that it is not the performance of language itself but a shared conceptual structure that should ground our determination of mutual understanding, and it is this very conceptual structure that we must aim to align between humans and AI models like ChatGPT.


\section{How do humans learn concepts?}

In cognitive science, as in philosophy, language and concepts are deeply intertwined. Human concept learning is a frequently-debated subject in cognitive science and developmental psychology, and there are many open questions about how human children acquire and understand novel concepts. One prominent theory, proposed by the developmental psychologist Susan Carey and known as Quinian bootstrapping \cite{susan2009origin}, posits that words are symbolic placeholders that children first learn and which gradually fill with meaning over time. Carey argues that the placeholder words are filled with meaning primarily through analogical reasoning \cite{susan2009origin}, while others have argued that they may be filled through a variety of other mechanisms \cite{jakab2013improve, beck2017can}.

Human concept learning is therefore also inextricably linked to the symbol grounding problem \cite{harnad1990symbol, searle1980minds}, identified in cognitive science, philosophy, and robotics as the challenge of linking symbols (like words) to the real-world phenomena they represent. 

If we think of words as placeholders for conceptual knowledge, veritable buckets to be filled by sensory and higher-level experiences (like thoughts, feelings, and memories) that define a concept, then the concept alignment problem becomes suddenly more concrete and accessible. Although two people probably have slightly different sensory information tied to the word ``apple'' (you may immediately think of a red, sweet one while I think of a green, tart one), their understanding of it is aligned enough to be able to communicate effectively about it \cite{lupyan-2023}. In their discourse they can safely trust that their companion understands what they mean by the word ``apple'' because the concept is grounded by similar sensorimotor, socio-cultural, and linguistic experiences on both sides. How do we achieve this for AI? 

Even between humans, grounding is often insufficient to ensure that we have aligned notions of a complex concept, like \textit{love} or \textit{democracy}, for example. But even in those cases, grounded language is a good starting point for concept-level alignment, because it allows us to discuss and come to a clearer understanding. Shared experience and shared grounded language forms the basis of our collective ability to understand the same concepts as someone else and make certain assumptions about what someone means when they use a word. 

Can we make those same assumptions about AI? Not yet, but we argue that we can get there by leveraging empirical tools already being developed in cognitive science and AI research.


\section{How do machines learn concepts?}

Probing machine understanding of concepts has been the domain of subfields like interpretability, multimodal learning, and language grounding \cite{tsai2019multimodal, doshi2017towards}. Each of these fields have developed methods of measuring to what extent words or labels are successfully linked to the real-world phenomena they describe. 

Neural networks form distributed representations, and it has proven difficult to map those to human-understandable concepts. In the context of deep neural networks, these ``representations'' are usually vectors of learned weights and biases in high-dimensional latent spaces. Researchers often measure ``representational alignment'' between multiple models or models and humans by using tools such as representational similarity analysis \cite{kriegeskorte2008representational} and centered kernel alignment \cite{kornblith2019similarity} which compare the representational geometries of two neural systems. Much of the literature in deep representation learning focuses on improving model performance by increasing the representational alignment of a student model with a skilled teacher (which can be a human or another model) using methods such as contrastive learning \cite{simclr} and knowledge distillation \cite{hinton2015distilling}. 
For example, analyzing representations might help us discover if the vector pairs for ``man''-``king'' and ``woman''-``queen'' are similarly positioned relative to each other in latent space \cite{vylomova2015take}. This representational analysis would suggest that a language model may have recovered some degree of human-like conceptual structure about the phenomena those words represent. However, while such representational correspondences provide clues to what is going on at the conceptual level, they do not guarantee that models are actually using human-understandable concepts in a human-like way \cite{peterson2020parallelograms}. Analyzing representations can, however, give us insight into how a neural network solves a specific problem, and help us begin to understand the relationship between representational alignment and concept alignment. 

Within interpretability, which started with simple tools like saliency maps for convolutional neural networks in computer vision \cite{simonyan2013deep, adebayo2018sanity}, the goal has always been to characterize machine representations or outputs in human-understandable terms. Saliency maps show which pixels of the original image are the most salient in the classification decision made by the model,  thus linking the visual input modality to the textual modality of the classification label. For years, adversarial machine learning was the prime example of machine intelligence gone wrong, where changing a few imperceptible (to humans) pixels could completely fool an AI model into miscategorizing an image \cite{kurakin2016adversarial}. Slowly, tools were developed to make models robust against such attacks. Interpretability work showed that robustified classifiers had more interpretable saliency maps \cite{tsipras2018robustness, etmann2019connection}, suggesting that having a more human-like (interpretable) conceptual representation meant improved model performance and protection from adversarial attacks. Other interpretability work has made the importance of concept-level alignment even more clear by showing that concept-level interpretability methods are simply more effective than pixel-based interpretability methods. There are even developed tools like Concept Activation Vectors \cite{kim2018interpretability} to probe a model's conceptual understanding.

The trend towards generalist models operating in many modalities \cite{radford2021learning, ramesh2021zero, driess2023palm} in recent years is beneficial to concept alignment research, because concept alignment is fundmentally multimodal. Many generalist, multimodal models have already provided proof by strong performance that a version of language-based bootstrapping is also incredibly effective for training deep multimodal AI models. Joint vision and language models like CLIP, DALL-E, and Imagen \cite{saharia2022photorealistic}, are already providing visual grounding to large text encoders. This brings us one step closer to shared grounded language with AI, because it allows for a kind of sensory meaning of what an AI system means when it says ``apple.'' This type of proof, grounded in a sensory modality like vision, is a first step towards human-like grounding. 

For example, the generative model Imagen, which takes in text prompts and generates images to match, uses a fixed T5 encoder as the LLM component, and then learns to map the correct visual information to an already rich textual embedding space of words. Other models like Parti \cite{yu2022scaling} start with a pretrained text encoder but then also finetune the LLM component further when trained jointly with visual input. There is an analogy here to Quinian bootstrapping, where the words are placeholders (although not empty ones, when they have pretrained text embeddings), and the meaning of the words is slowly filled and enriched through information from the other modalities, in this case through vision. 


\section{How do humans align on concepts?}

Concept alignment between humans is not a static process. Shared, conceptually-grounded language gives human agents the ability to make reasonable assumptions about each others' intended meanings. Humans further hone their concept alignment through interaction. While language is perhaps the most important way that humans align on concepts, there are important subtleties in this process occurring under the surface of verbal interactions between human agents.

The past several decades has seen the emergence of behavioral and neuroscientific evidence in cognitive science for the role of interaction in human communication and comprehension \cite{paxton-2016, wheatley-2019}. A cluster of effects that have been studied throughout the years, such as ``behavioral matching,'' ``imitation,'' ``mimicry,'' ``entrainment,'' ``repetition,'' ``accommodation,'' ``coordination,'' and ``joint-action,'' is currently coalescing into a single, coherent cognitive phenomenon referred to in the cognitive science literature as ``alignment'' \cite{rasenberg-2020}.

The term ``alignment'' in this context is drawn from the interactive alignment paradigm \cite{pickering-2004}, an influential theory in cognitive science which argues that successful human communication is facilitated by aligning linguistic processing across multiple scales.\footnote{Lexical entrainment (that interlocutors will use or repeat similar words throughout the course of a conversation, \cite{brennan-1996conceptualpacts, brennan-1996lexical}) and syntactic priming (that interlocutors will use or repeat similar syntactic structures throughout the course of a conversation, \cite{pickering-1999}) are key examples of alignment in this sense.} This approach has since been extended by the analysis of human concept alignment \textit{qua} interpersonal synergies, an analysis of communicative coordination using dynamical systems theory \cite{louwerse-2012, dale-2013, fusaroli-2013, duran-2014}. Interactive adaptation of this kind has been shown to be crucial for resolving misunderstanding and more accurately fine-tuning aligned conceptual understandings during real-time problem solving and coordination \cite{fusaroli-2014}.

Drawing on these paradigms, a range of measures have been developed in cognitive science for measuring human alignment. For example, Abney et al. (2015) develop a measure of synergy in coordination that predicts performance in a cooperative task during problem solving \cite{abney-2015}. This finding indicates that measures of synergy might be useful for evaluating concept alignment between humans and AI.

Additionally, maximizing information exchange between two complex systems (i.e., increasing the ability for the dynamics of each system to affect, and be affected by, the other, also known as ``resonance'' \cite{west-2008}) may be a key driver of concept alignment. Complexity matching, a measure of such information exchange,\footnote{Complexity matching is measured using time series data, and thus may be implemented in any modality from which a time series can be extracted.} has been shown to increase between participants in an affiliative conversation and decrease between partners in an argumentative conversation \cite{abney-2014}. This provides empirical evidence for a connection between information exchange and concept alignment, making measures such as complexity matching another promising measure of concept alignment.

We argue that in addition to grounded linguistic meaning, human-AI alignment must be dynamic and adaptive in order to fine-tune shared conceptual understanding. Interaction serves to align diverse experiences and perspectives between agents and create paths for shared meaning to take hold. Creating AI systems that are capable of engaging in and reciprocating this sort of multiscale, interactive adaptation with their interlocutors will be an important step towards human-aligned AI that reliably meets the standard for concept alignment that we would reasonably expect of a human agent.


\section{How might humans and machines align on concepts?}

It is clear that the standard for proof of human-AI alignment must be even higher than it is for human-human alignment, because we can make more assumptions about another human's understanding of a shared word than we can about, say, ChatGPT. We have more reason to believe an AI might be misaligned due to the lack of this shared embodiment, shared experience, and shared linguistic grounding.

To develop a standard for measuring, steering, and improving concept alignment, we must start with the seeds laid already within the AI research space, often under areas such as interpretability, language grounding, grounded robotics, and multimodal models. 

Promisingly, it is clear that bootstrapping from LLM representations and then grounding in other modalities is a powerful combination for AI systems. Newer robotics models like PaLM-E \cite{driess2023palm} start with LLM representations and then enrich those with sensory information that the robot experiences. The sensory information improves the word representations, and the rich existing knowledge embedded in the LLM helps the robot bootstrap instead of relying solely on sensorimotor experience as a source of all its knowledge. This method is extremely successful, and PaLM-E achieved state-of-the-art performance on many robot learning tasks. Interestingly, PaLM-E also became an excellent vision-and-language model, apparently benefiting from the addition of sensorimotor modalities to improve its performance in the vision-language space. This is a first step towards the kind of concept alignment that we need -- first bootstrapping using pretrained LLMs and then filling words with conceptual meaning grounded in other modalities should lead to more verifiably human-like concepts and better performance. Different modalities provide ways for an AI system to demonstrate, through an image or a sound, that it \textit{understands} a concept in a human-like way. 

There is clearly still work to be done in designing AI systems that are interpretable and grounded at the concept level. Further inquiry along these lines is critical to achieve concept alignment. Technical tools for probing the effects of language indeterminacy must be developed to measure, stress-test, and validate claims of concept alignment.

We argue that this work should be done in conversation with and guided by the cognitive science research in concept formation and alignment. What degree of shared conceptual understanding is required, or even commonly present, between interacting humans is something we can only fully characterize through empirical studies of human agents. We can then design AI to have, at a lower bound, this degree of shared conceptual understanding. This is a complex and iterative process requiring ongoing studies and engineering across modalities.

For example, say we determine through child development studies that visual recognition of an ``apple'' is usually learned in the first 18 months of development, that sensorimotor knowledge of holding, biting, and tasting an apple is usually learned within the first 36 months, and that reliably using the word correctly in a linguistic context is learned within the first five years. In addition, we may find that most adult humans think of a red, sweet apple when they think of apples, but that a large proportion also think of Granny Smiths, and a smaller proportion of yellow delicious apples. In cognitive science research, prototype theory \cite{rosch1973natural} has helped us extensively explore such ideas about the prototypicality of particular instantiations of a concept -- whether a red apple is more prototypical than a green one, for example. Cognitive science helps us set a standard for what a fellow human might assume our shared understanding of the concept ``apple'' would be, a standard which can then be applied to machines. This standard will change for more complex concepts, but it will always be grounded in what can be reasonably expected of humans. This isn't a perfect process of course; for example, a fellow human who hails from a geographic region without access to apples might lack this shared understanding, and they would still be able to interactively communicate this, which is why the interactive and adaptive component is so important. However, aiming for shared understanding in a way that approximates shared understanding between humans is a good place to start. The beauty of concept alignment is that it allows us to reliably set goals for AI systems and tractably measure performance against these goals, because concepts are intrinsically human-understandable and can be tractably mapped to standards expected of a reasonable human agent. The standards for concept understanding and alignment expected of a reasonable human agent should continue to be characterized by a vital ongoing dialogue between cognitive science research and AI research.

A human agent's understanding of concepts is also fine-tuned through interaction \cite{mandler2004foundations}, and humans can even use different understandings of the same concept based on context (for example, if I explain to you that I'm talking about a green apple, it's easy for you to make the switch even if you were thinking about a red one before). It is important that we also get AI systems to the stage of concept alignment where they can reliably interact with humans to fine-tune their conceptual understandings as a reasonable human would. Fine-tuning through interaction can be informed by work in human-in-the-loop-learning (HILL) in the machine learning literature \cite{li2017dialogue}, and work in cognitive alignment in the cognitive science literature \cite{murthy2022shades, hawkins2023partners}. Among AI researchers, one popular way of approaching interaction currently is reinforcement learning from human feedback (RLHF) \cite{ziegler2019fine, ouyang2022training}, which asks human participants to rank instances of an AI agent's behavior and uses this feedback to train a reward model to optimize the AI agent's policy. However, RLHF currently operates at the behavioral level. We don't know exactly how policy changes resulting from the human feedback filter through to the AI systems' understanding of concepts, and specifically whether RLHF in its current form encourages the AI to form human-like concepts at all. RLHF is an interesting start, particularly because it incorporates human preferences through interaction. However, more work along these lines, focused at the concept-level, is necessary to interactively improve concept alignment through human feedback. Verifying and improving alignment through conversation and nonverbal cues is critical to successful human collaboration. Concept alignment between humans and AI would be incomplete without this capacity for interactive fine-tuning and improvement.




\section{Conclusion: Where do we go from here?}

We have introduced the idea of concept alignment between humans and AI, explained why it is an important prerequisite to discussions of value alignment, and charted a path forward towards concept alignment. In particular, we argue that: 
\begin{itemize}
    \item Concept alignment across sensory modalities that are part of the human experience will lead to more human-aligned natural language representations (vision, textual language, speech/audio, olfaction, touch, vestibular and limbic systems, etc.)
    
    \item Language is a vital, but not sufficient, part of concept alignment. \\
    Joint training in generalist models can use concept alignment verification as a feedback loop for improving and grounding natural language representations, aligning them with shared meaning expected of a reasonable human agent.
    
    \item Concept alignment will lead to natural language communication with AI systems that is functionally equivalent to natural language communication among humans.
\end{itemize}
Concept alignment, and the shared grounded language that will emerge from it, will lay the groundwork for value alignment, behavior alignment, and other important subgoals that are an integral step on the journey towards AI alignment. 


\section*{Acknowledgements}
The authors would like to thank the Diverse Intelligences Summer Institute (DISI) and funding from the Templeton World Charity Foundation (https://doi.org/10.54224/20333) for making this collaboration possible.



\bibliography{main}

\begin{thebibliography}{10}

\bibitem{chatgpt}
\url{https://openai.com/blog/chatgpt/}, 2022.
\newblock ChatGPT, OpenAI.

\bibitem{abney-2014}
Drew~H Abney, Alexandra Paxton, Rick Dale, and Christopher~T Kello.
\newblock Complexity matching in dyadic conversation.
\newblock {\em Journal of Experimental Psychology: General}, 143(6):2304, 2014.

\bibitem{abney-2015}
Drew~H Abney, Alexandra Paxton, Rick Dale, and Christopher~T Kello.
\newblock Movement dynamics reflect a functional role for weak coupling and role structure in dyadic problem solving.
\newblock {\em Cognitive processing}, 16:325--332, 2015.

\bibitem{adebayo2018sanity}
Julius Adebayo, Justin Gilmer, Michael Muelly, Ian Goodfellow, Moritz Hardt, and Been Kim.
\newblock Sanity checks for saliency maps.
\newblock {\em Advances in neural information processing systems}, 31, 2018.

\bibitem{beck2017can}
Jacob Beck.
\newblock Can bootstrapping explain concept learning?
\newblock {\em Cognition}, 158:110--121, 2017.

\bibitem{BonJour-1985}
Laurence BonJour.
\newblock {\em The Structure of Empirical Knowledge}.
\newblock Harvard University Press, Cambridge, Mass., 1985.

\bibitem{brennan-1996conceptualpacts}
Susan~E Brennan and Herbert~H Clark.
\newblock Conceptual pacts and lexical choice in conversation.
\newblock {\em Journal of experimental psychology: Learning, memory, and cognition}, 22(6):1482, 1996.

\bibitem{brennan-1996lexical}
Susan~E Brennan et~al.
\newblock Lexical entrainment in spontaneous dialog.
\newblock {\em Proceedings of ISSD}, 96:41--44, 1996.

\bibitem{brown2021value}
Daniel~S Brown, Jordan Schneider, Anca Dragan, and Scott Niekum.
\newblock Value alignment verification.
\newblock In {\em International Conference on Machine Learning}, pages 1105--1115. PMLR, 2021.

\bibitem{susan2009origin}
Susan Carey.
\newblock {\em The Origin of Concepts}.
\newblock Oxford Series in Cognitive Development. Oxford University Press, USA, 2009.

\bibitem{simclr}
Ting Chen, Simon Kornblith, Mohammad Norouzi, and Geoffrey Hinton.
\newblock A simple framework for contrastive learning of visual representations.
\newblock In {\em International conference on machine learning}, pages 1597--1607. PMLR, 2020.

\bibitem{christian2020alignment}
B.~Christian.
\newblock {\em The Alignment Problem: Machine Learning and Human Values}.
\newblock WW Norton, 2020.

\bibitem{dale-2013}
Rick Dale, Riccardo Fusaroli, Nicholas~D Duran, and Daniel~C Richardson.
\newblock The self-organization of human interaction.
\newblock In {\em Psychology of learning and motivation}, volume~59, pages 43--95. Elsevier, 2013.

\bibitem{doshi2017towards}
Finale Doshi-Velez and Been Kim.
\newblock Towards a rigorous science of interpretable machine learning.
\newblock {\em arXiv preprint arXiv:1702.08608}, 2017.

\bibitem{driess2023palm}
Danny Driess, Fei Xia, Mehdi~SM Sajjadi, Corey Lynch, Aakanksha Chowdhery, Brian Ichter, Ayzaan Wahid, Jonathan Tompson, Quan Vuong, Tianhe Yu, et~al.
\newblock Palm-e: An embodied multimodal language model.
\newblock {\em arXiv preprint arXiv:2303.03378}, 2023.

\bibitem{duran-2014}
Nicholas~D Duran and Rick Dale.
\newblock Perspective-taking in dialogue as self-organization under social constraints.
\newblock {\em New Ideas in Psychology}, 32:131--146, 2014.

\bibitem{etmann2019connection}
Christian Etmann, Sebastian Lunz, Peter Maass, and Carola-Bibiane Sch{\"o}nlieb.
\newblock On the connection between adversarial robustness and saliency map interpretability.
\newblock {\em arXiv preprint arXiv:1905.04172}, 2019.

\bibitem{fusaroli-2013}
Riccardo Fusaroli, Drew~H Abney, Bahador Bahrami, Christopher~T Kello, and Kristian Tyl{\'e}n.
\newblock Conversation, coupling and complexity: Matching scaling laws predict performance in a joint decision task.
\newblock In {\em Poster presented at the 35th annual conference of the cognitive science society}, 2013.

\bibitem{fusaroli-2014}
Riccardo Fusaroli, Joanna R{\k{a}}czaszek-Leonardi, and Kristian Tyl{\'e}n.
\newblock Dialog as interpersonal synergy.
\newblock {\em New Ideas in Psychology}, 32:147--157, 2014.

\bibitem{gabriel2020artificial}
Iason Gabriel.
\newblock Artificial intelligence, values, and alignment.
\newblock {\em Minds and machines}, 30(3):411--437, 2020.

\bibitem{uber}
Richard Gonzales.
\newblock Feds say self-driving uber suv did not recognize jaywalking pedestrian in fatal crash.
\newblock \url{https://www.npr.org/2019/11/07/777438412/feds-say-self-driving-uber-suv-did-not-recognize-jaywalking-pedestrian}, 2019.
\newblock National Public Radio (NPR).

\bibitem{hadfield2017inverse}
Dylan Hadfield-Menell, Smitha Milli, Pieter Abbeel, Stuart~J Russell, and Anca Dragan.
\newblock Inverse reward design.
\newblock {\em Advances in neural information processing systems}, 30, 2017.

\bibitem{hadfield2016cooperative}
Dylan Hadfield-Menell, Stuart~J Russell, Pieter Abbeel, and Anca Dragan.
\newblock Cooperative inverse reinforcement learning.
\newblock {\em Advances in neural information processing systems}, 29, 2016.

\bibitem{harnad1990symbol}
Stevan Harnad.
\newblock The symbol grounding problem.
\newblock {\em Physica D: Nonlinear Phenomena}, 42(1-3):335--346, 1990.

\bibitem{hawkins2023partners}
Robert~D Hawkins, Michael Franke, Michael~C Frank, Adele~E Goldberg, Kenny Smith, Thomas~L Griffiths, and Noah~D Goodman.
\newblock From partners to populations: A hierarchical bayesian account of coordination and convention.
\newblock {\em Psychological Review}, 130(4):977, 2023.

\bibitem{hendrycks2020aligning}
Dan Hendrycks, Collin Burns, Steven Basart, Andrew Critch, Jerry Li, Dawn Song, and Jacob Steinhardt.
\newblock Aligning ai with shared human values.
\newblock {\em arXiv preprint arXiv:2008.02275}, 2020.

\bibitem{hinton2015distilling}
Geoffrey Hinton, Oriol Vinyals, and Jeff Dean.
\newblock Distilling the knowledge in a neural network.
\newblock {\em arXiv preprint arXiv:1503.02531}, 2015.

\bibitem{newyorker}
Matthew Hutson.
\newblock Can we stop runaway a.i.?
\newblock \url{https://www.newyorker.com/science/annals-of-artificial-intelligence/can-we-stop-the-singularity}, 2023.
\newblock The New Yorker.

\bibitem{jakab2013improve}
Zoltan Jakab.
\newblock How to improve on quinian bootstrapping-a response to nativist objections.
\newblock In {\em Proceedings of the Annual Meeting of the Cognitive Science Society}, volume~35, 2013.

\bibitem{kim2018interpretability}
Been Kim, Martin Wattenberg, Justin Gilmer, Carrie Cai, James Wexler, Fernanda Viegas, et~al.
\newblock Interpretability beyond feature attribution: Quantitative testing with concept activation vectors (tcav).
\newblock In {\em International conference on machine learning}, pages 2668--2677. PMLR, 2018.

\bibitem{kornblith2019similarity}
Simon Kornblith, Mohammad Norouzi, Honglak Lee, and Geoffrey Hinton.
\newblock Similarity of neural network representations revisited.
\newblock In {\em International conference on machine learning}, pages 3519--3529. PMLR, 2019.

\bibitem{kriegeskorte2008representational}
Nikolaus Kriegeskorte, Marieke Mur, and Peter~A Bandettini.
\newblock Representational similarity analysis-connecting the branches of systems neuroscience.
\newblock {\em Frontiers in systems neuroscience}, page~4, 2008.

\bibitem{Kuhn-1978}
Thomas~S. Kuhn.
\newblock The essential tension.
\newblock {\em Philosophy of Science}, 45(4):649--652, 1978.

\bibitem{kurakin2016adversarial}
Alexey Kurakin, Ian Goodfellow, and Samy Bengio.
\newblock Adversarial machine learning at scale.
\newblock {\em arXiv preprint arXiv:1611.01236}, 2016.

\bibitem{li2017dialogue}
Jiwei Li, Alexander~H Miller, Sumit Chopra, Marc'Aurelio Ranzato, and Jason Weston.
\newblock Dialogue learning with human-in-the-loop.
\newblock In {\em 5th International Conference on Learning Representations, ICLR 2017}, 2017.

\bibitem{louwerse-2012}
Max~M Louwerse, Rick Dale, Ellen~G Bard, and Patrick Jeuniaux.
\newblock Behavior matching in multimodal communication is synchronized.
\newblock {\em Cognitive science}, 36(8):1404--1426, 2012.

\bibitem{lupyan-2023}
Gary Lupyan, Ryutaro Uchiyama, Bill Thompson, and Daniel Casasanto.
\newblock Hidden differences in phenomenal experience.
\newblock {\em Cognitive Science}, 47(1):e13239, 2023.

\bibitem{raceai}
Ryan Mac.
\newblock Facebook apologizes after a.i. puts ‘primates’ label on video of black men.
\newblock \url{https://www.nytimes.com/2021/09/03/technology/facebook-ai-race-primates.html}, 2021.
\newblock The New York Times.

\bibitem{mandler2004foundations}
Jean~Matter Mandler.
\newblock {\em The foundations of mind: Origins of conceptual thought}.
\newblock Oxford University Press, 2004.

\bibitem{alphago-wired}
Cade Metz.
\newblock How google's ai viewed the move no human could understand.
\newblock \url{https://www.wired.com/2016/03/googles-ai-viewed-move-no-human-understand/}, 2016.
\newblock WIRED.

\bibitem{murthy2022shades}
Sonia~K Murthy, Thomas~L Griffiths, and Robert~D Hawkins.
\newblock Shades of confusion: Lexical uncertainty modulates ad hoc coordination in an interactive communication task.
\newblock {\em Cognition}, 225:105152, 2022.

\bibitem{Neurath-1983}
Otto Neurath.
\newblock {\em Protocol Statements}, page 91–99.
\newblock Springer Netherlands, Dordrecht, 1983.

\bibitem{ouyang2022training}
Long Ouyang, Jeffrey Wu, Xu~Jiang, Diogo Almeida, Carroll Wainwright, Pamela Mishkin, Chong Zhang, Sandhini Agarwal, Katarina Slama, Alex Ray, et~al.
\newblock Training language models to follow instructions with human feedback.
\newblock {\em Advances in Neural Information Processing Systems}, 35:27730--27744, 2022.

\bibitem{paxton-2016}
Alexandra Paxton, Rick Dale, and Daniel~C Richardson.
\newblock Social coordination of verbal and nonverbal behaviours.
\newblock {\em Interpersonal coordination and performance in social systems}, page 259, 2016.

\bibitem{peterson2020parallelograms}
JC~Peterson, D~Chen, and TL~Griffiths.
\newblock Parallelograms revisited: Exploring the limitations of vector space models for simple analogies. cognition, 205, article 104440, 2020.

\bibitem{piaget-1967}
J~Piaget and B~Inhelder.
\newblock Systems of reference and horizontal--vertical coordinates.
\newblock {\em The Child’s Conception of Space (1967)}, pages 375--418, 1967.

\bibitem{pickering-1999}
Martin~J Pickering and Holly~P Branigan.
\newblock Syntactic priming in language production.
\newblock {\em Trends in cognitive sciences}, 3(4):136--141, 1999.

\bibitem{pickering-2004}
Martin~J Pickering and Simon Garrod.
\newblock Toward a mechanistic psychology of dialogue.
\newblock {\em Behavioral and brain sciences}, 27(2):169--190, 2004.

\bibitem{Quine-1951}
Willard V.~O. Quine.
\newblock Two dogmas of empiricism.
\newblock {\em Philosophical Review}, 60(1):20--43, 1951.

\bibitem{Quine-1968}
W.V.O. Quine.
\newblock Ontological relativity.
\newblock {\em Journal of Philosophy}, 65(7):185--212, 1968.

\bibitem{radford2021learning}
Alec Radford, Jong~Wook Kim, Chris Hallacy, Aditya Ramesh, Gabriel Goh, Sandhini Agarwal, Girish Sastry, Amanda Askell, Pamela Mishkin, Jack Clark, et~al.
\newblock Learning transferable visual models from natural language supervision.
\newblock In {\em International conference on machine learning}, pages 8748--8763. PMLR, 2021.

\bibitem{ramesh2021zero}
Aditya Ramesh, Mikhail Pavlov, Gabriel Goh, Scott Gray, Chelsea Voss, Alec Radford, Mark Chen, and Ilya Sutskever.
\newblock Zero-shot text-to-image generation.
\newblock In {\em International Conference on Machine Learning}, pages 8821--8831. PMLR, 2021.

\bibitem{rasenberg-2020}
Marlou Rasenberg, Asli {\"O}zy{\"u}rek, and Mark Dingemanse.
\newblock Alignment in multimodal interaction: An integrative framework.
\newblock {\em Cognitive science}, 44(11):e12911, 2020.

\bibitem{ribeiro2016should}
Marco~Tulio Ribeiro, Sameer Singh, and Carlos Guestrin.
\newblock " why should i trust you?" explaining the predictions of any classifier.
\newblock In {\em Proceedings of the 22nd ACM SIGKDD international conference on knowledge discovery and data mining}, pages 1135--1144, 2016.

\bibitem{rosch1973natural}
Eleanor~H Rosch.
\newblock Natural categories.
\newblock {\em Cognitive psychology}, 4(3):328--350, 1973.

\bibitem{saharia2022photorealistic}
Chitwan Saharia, William Chan, Saurabh Saxena, Lala Li, Jay Whang, Emily~L Denton, Kamyar Ghasemipour, Raphael Gontijo~Lopes, Burcu Karagol~Ayan, Tim Salimans, et~al.
\newblock Photorealistic text-to-image diffusion models with deep language understanding.
\newblock {\em Advances in Neural Information Processing Systems}, 35:36479--36494, 2022.

\bibitem{searle1980minds}
John~R Searle.
\newblock Minds, brains, and programs.
\newblock {\em Behavioral and brain sciences}, 3(3):417--424, 1980.

\bibitem{silver2016mastering}
David Silver, Aja Huang, Chris~J Maddison, Arthur Guez, Laurent Sifre, George Van Den~Driessche, Julian Schrittwieser, Ioannis Antonoglou, Veda Panneershelvam, Marc Lanctot, et~al.
\newblock Mastering the game of go with deep neural networks and tree search.
\newblock {\em nature}, 529(7587):484--489, 2016.

\bibitem{simonyan2013deep}
K~Simonyan, A~Vedaldi, and A~Zisserman.
\newblock Deep inside convolutional networks: visualising image classification models and saliency maps.
\newblock In {\em Proceedings of the International Conference on Learning Representations (ICLR)}. ICLR, 2014.

\bibitem{su}
SingularityGroup.
\newblock from risk to reward: the role of ai alignment in shaping a positive future.
\newblock \url{https://www.su.org/blog/from-risk-to-reward-the-role-of-ai-alignment-in-shaping-a-positive-future}, 2023.

\bibitem{Taylor-1982}
Charles Taylor.
\newblock Rationality.
\newblock In Martin Hollis and Steven Lukes, editors, {\em Rationality and Relativism}, pages 87--105. MIT Press, 1982.

\bibitem{tsai2019multimodal}
Yao-Hung~Hubert Tsai, Shaojie Bai, Paul~Pu Liang, J~Zico Kolter, Louis-Philippe Morency, and Ruslan Salakhutdinov.
\newblock Multimodal transformer for unaligned multimodal language sequences.
\newblock In {\em Proceedings of the conference. Association for Computational Linguistics. Meeting}, volume 2019, page 6558. NIH Public Access, 2019.

\bibitem{tsipras2018robustness}
Dimitris Tsipras, Shibani Santurkar, Logan Engstrom, Alexander Turner, and Aleksander Madry.
\newblock Robustness may be at odds with accuracy.
\newblock {\em arXiv preprint arXiv:1805.12152}, 2018.

\bibitem{vylomova2015take}
Ekaterina Vylomova, Laura Rimell, Trevor Cohn, and Timothy Baldwin.
\newblock Take and took, gaggle and goose, book and read: Evaluating the utility of vector differences for lexical relation learning.
\newblock {\em arXiv preprint arXiv:1509.01692}, 2015.

\bibitem{west-2008}
Bruce~J West, Elvis~L Geneston, and Paolo Grigolini.
\newblock Maximizing information exchange between complex networks.
\newblock {\em Physics Reports}, 468(1-3):1--99, 2008.

\bibitem{wheatley-2019}
Thalia Wheatley, Adam Boncz, Ivan Toni, and Arjen Stolk.
\newblock Beyond the isolated brain: The promise and challenge of interacting minds.
\newblock {\em Neuron}, 103(2):186--188, 2019.

\bibitem{yu2022scaling}
Jiahui Yu, Yuanzhong Xu, Jing~Yu Koh, Thang Luong, Gunjan Baid, Zirui Wang, Vijay Vasudevan, Alexander Ku, Yinfei Yang, Burcu~Karagol Ayan, et~al.
\newblock Scaling autoregressive models for content-rich text-to-image generation.
\newblock {\em arXiv preprint arXiv:2206.10789}, 2(3):5, 2022.

\bibitem{zhong2023clock}
Ziqian Zhong, Ziming Liu, Max Tegmark, and Jacob Andreas.
\newblock The clock and the pizza: Two stories in mechanistic explanation of neural networks.
\newblock {\em arXiv preprint arXiv:2306.17844}, 2023.

\bibitem{ziegler2019fine}
Daniel~M Ziegler, Nisan Stiennon, Jeffrey Wu, Tom~B Brown, Alec Radford, Dario Amodei, Paul Christiano, and Geoffrey Irving.
\newblock Fine-tuning language models from human preferences.
\newblock {\em arXiv preprint arXiv:1909.08593}, 2019.

\end{thebibliography}
\bibliographystyle{plain}


\end{document}